\documentclass{article}

\usepackage[final]{corl_2020}

\usepackage{etoolbox}
\providetoggle{final}
\settoggle{final}{true}

\usepackage[pdftex]{graphicx}
\usepackage{amsmath}
\usepackage{amssymb}
\usepackage{marvosym}
\usepackage{dblfloatfix}
\usepackage{floatrow}

\DeclareFixedFont{\ttb}{T1}{txtt}{bx}{n}{7} % for bold
\DeclareFixedFont{\ttm}{T1}{txtt}{m}{n}{7}  % for normal
\usepackage{color}
\definecolor{deepblue}{rgb}{0,0,0.5}
\definecolor{deepred}{rgb}{0.6,0,0}
\definecolor{deepgreen}{rgb}{0,0.5,0}
\definecolor{deepgrey}{rgb}{0.6,0.6,0.6}
\usepackage{listings}
\newcommand\pythonstyle{\lstset{
  language=Python,
  basicstyle=\small\ttm,
  keywordstyle=\ttb\color{deepblue},
  keywords={from, import, self, fit, make},             % Add keywords here
  emph={MyClass,__init__},          % Custom highlighting
  emphstyle=\ttb\color{deepred},    % Custom highlighting style
  stringstyle=\color{deepgreen},
  commentstyle=\ttm\color{deepgrey},
  frame=tb,                         % Any extra options here
  showstringspaces=false,
  captionpos=b
}}
\lstnewenvironment{python}[1][]{
  \pythonstyle
  \lstset{caption=#1}
}{}
\newcommand\pythoninline[1]{{\pythonstyle\lstinline!#1!}}

\title{\iftoggle{final}{OffWorld}{(Anonymized)} Gym: open-access physical robotics environment for real-world reinforcement learning benchmark and research}

\author{
  Ashish Kumar$^{\text{\Yinyang}}$ \\
  \And
  Toby Buckley\phantom{12}\  \\
  \And
  John B. Lanier\phantom{123}\  \\
  \AND
  Qiaozhi Wang \\
  \And
  Alicia Kavelaars \\
  \And
  Ilya Kuzovkin$^{\text{\Yinyang},\text{\Letter}}$
  \AND
  \\
  \textsuperscript{\textsc{OFF}}\textsc{world} \\
  Pasadena, CA, USA
  \AND
  \\
  $^{\text{\Letter}}\ $Corresponding author: \texttt{ilya.kuzovkin@offworld.ai}
}

\begin{document}

\maketitle
\thispagestyle{empty}
\pagestyle{empty}

%
%  Abstract
%
\begin{abstract}
Success stories of applied machine learning can be traced back to the datasets and environments that were put forward as challenges for the community. The challenge that the community sets as a benchmark is usually the challenge that the community eventually solves. The ultimate challenge of reinforcement learning research is to train \emph{real} agents to operate in the \emph{real} environment, but there is no common real-world benchmark to track the progress of RL on physical robotic systems. To address this issue we have created \iftoggle{final}{OffWorld}{(Anonymized)} Gym -- a collection of real-world environments for reinforcement learning in robotics with free public remote access. In this work, we introduce four tasks in two environments and experimental results on one of them that demonstrate the feasibility of learning on a real robotic system. We train a mobile robot end-to-end to solve simple navigation task relying solely on camera input and without the access to location information. Close integration into existing ecosystem allows the community to start using \iftoggle{final}{OffWorld}{(Anonymized)} Gym without any prior experience in robotics and takes away the burden of managing a physical robotics system, abstracting it under a familiar API. To start training, visit \iftoggle{final}{\texttt{\url{https://gym.offworld.ai}}}{\texttt{\url{https://anonymized}} \emph{(please find the anonymized screenshots of the critical components of the online system attached to this submission)}}.
\end{abstract}

%
%  Introduction
%
\section{INTRODUCTION}
Reinforcement learning~\cite{sutton2018reinforcement} offers a strong framework to approach machine learning problems that can be formulated in terms of \emph{agents} operating in \emph{environments} and receiving \emph{rewards}. Coupled with the representational power and capacity of deep neural networks~\cite{lecun2015deep}, this framework has enabled artificial agents to achieve superhuman performance in Atari games~\cite{mnih2015human}, Go~\cite{silver2017mastering}, and real time strategy games such as Dota~2~\cite{OpenAI_dota} and StarCraft II~\cite{alphastarblog}. Deep reinforcement learning has been successfully applied to simulated environments, demonstrating the ability to solve control problems in discrete~\cite{hessel2018rainbow, mnih2016asynchronous, schulman2017proximal} and continuous~\cite{lillicrap2015continuous, duan2016benchmarking} action spaces, perform long-term planning~\cite{guo2014deep, guez2019investigation}, use memory~\cite{wayne2018unsupervised}, explore environments efficiently~\cite{ecoffet2019go}, and even learn to communicate with other agents~\cite{das2018tarmac}. These and many other capabilities proven by deep reinforcement learning (DRL) methods~\cite{li2018deep} hold an inspiring promise of the applicability of DRL to real world tasks, particularly in the field of robotics. 

Despite the fact that many consider operations in real world settings to be the ultimate challenge for reinforcement learning research~\cite{dulac2019challenges}, the search for solutions to that challenge is being carried out predominantly in simulated environments~\cite{heess2015memory, heess2015learning, schulman2015high, schulman2015trust, duan2016benchmarking, houthooft2016vime, gu2016continuous, mnih2016asynchronous, zhu2017target, hausman2018learning}. This focus on simulated environments as opposed to physical ones can be attributed to the high difficulty of training in real world environments. High sample complexity of modern DRL methods makes collecting a sufficient amount of observations on a real robotic system both time consuming and challenging from a maintenance standpoint. As a result, the training of real world agents has been approached in a variety of ways, both directly~\cite{zhang2015towards, lenz2015deepmpc, levine2016end, finn2016guided, yahya2017collective, levine2018learning, mahmood2018benchmarking, kalashnikov2018qt} and using simulation-to-real transfer learning to minimize experience needed in a real setting~\cite{rusu2016sim, tobin2017domain, andrychowicz2018learning}. Recent works on imitation learning~\cite{finn2016guided, duan2017one, finn2017one, christiano2017deep, nair2018overcoming, sun2019provably} and reduction of sample complexity~\cite{gupta2018meta, clavera2018model, pathak2019self, sahni2019addressing, vezzani2019learning} also provide a path towards making training in real feasible.

From the previous major successes of machine learning, we see that the goal the community sets as a benchmark is usually the goal that the community eventually solves. Thus to solve the hard problems in RL for the real world, the RL community must add real-world environments to their set of benchmarks. Adding a common physical benchmark environment to the set of canonical reference tasks such as Atari games~\cite{bellemare2013arcade} and MuJoCo creatures~\cite{todorov2012mujoco} would enable future research to take into account, and hopefully accelerate, the applicability of RL methods to real world robotics.

\begin{figure}[t]
    \includegraphics[width=\linewidth]{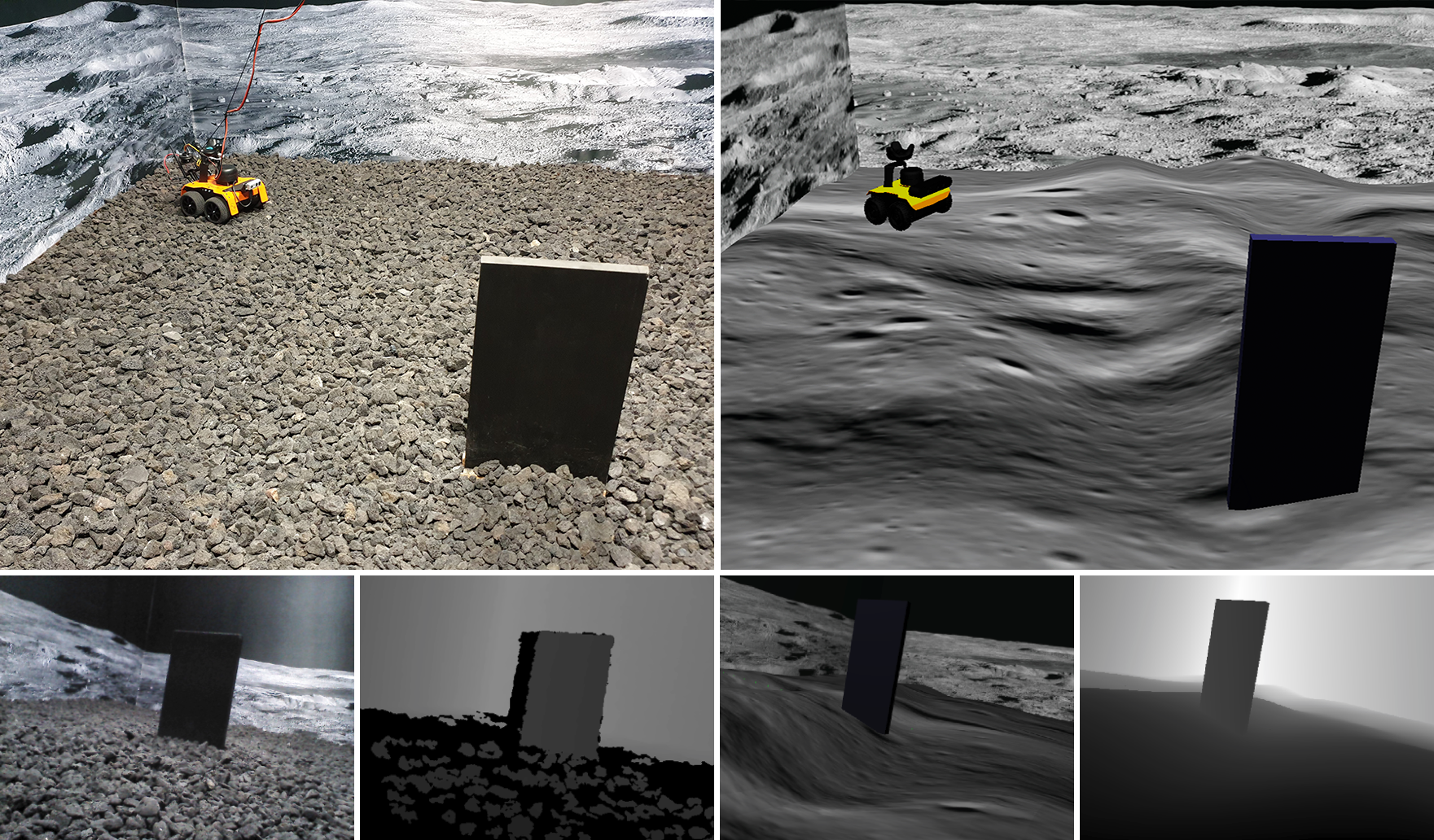}
    \caption{The top row shows the real (left) and the simulated (right) instances of the \texttt{MonolithDiscrete} environment. The users have same access to both via the same API interface, allowing for a seamless transition between a simulated and a real versions of an environment. The bottom row shows RGB and depth inputs in both instances from the robot's perspective.}
    \label{fig:sim-real-env}
\end{figure}

In this work, we present four real-world, publicly-accessible, remote-operated robotics RL environments from the \iftoggle{final}{OffWorld}{(Anonymized)} Gym framework\footnote{A video of an agent controlling a physical mobile robot in a real environment: \iftoggle{final}{\url{https://www.youtube.com/watch?v=lgwaZHxtJc0}}{\texttt{\url{https://youtu.be/FkUJ_Gl1hJU}} \emph{(anonymized)}}}, consisting of two-tasks in both discrete and continuous control formulations. These environments conform to the OpenAI \texttt{gym} API while remote-controlling a real robot maintained by the authors and address general robotics challenges such as locomotion, navigation, planning, and obstacle avoidance. In each task, the robot must reach a visual beacon while relying solely on visual input. In the first, the robot is situated on an open, uneven terrain, while in the second, the robot is surrounded by obstacles, which it must implicitly learn to navigate around. The learning methods that the research community will find to achieve robust performance in these tasks can then be naturally transferred to the corresponding applications in other real world robotics domains. Simulated variants of these environments are also provided.

\iftoggle{final}{OffWorld Inc.}{(Anonymized)} is committed to providing long-term free support and maintenance of the physical environments, as well as constructing additional ones to expand the set of challenges and meet the demand of the community.

%
%  Related work
%
\section{RELATED WORK}
\label{sec:related}
Publicly available simulated environments play an important role in the development of RL methods, providing a common ground for comparing different approaches and allowing progress in the field to be explicitly tracked. However, they do not allow to bridge the gap between simulation and reality. Simulated environments address various general aspects of reinforcement learning research such as control~\cite{bellemare2013arcade}, navigation~\cite{beattie2016deepmind, kempka2016vizdoom, johnson2016malmo, habitat19arxiv}, physical interactions~\cite{todorov2012mujoco} and perception~\cite{xia2018gibson}. More domain-specific simulated environments explore such fields as robotics~\cite{klimov2017roboschool, openai2018ingredients, zamora2016extending} and autonomous driving~\cite{dosovitskiy2017carla}. 

Following the signs of applicability of RL in real-world robotics, RL-oriented hardware kits became available in the past year to support the development of reproducible RL in robotics research~\cite{gealy2019quasi, yang2019replab}. Mandlekar at al.~\cite{mandlekar2018roboturk} and Orrb et al.~\cite{orrb2019} introduce platforms for generating high fidelity robot interaction data that can be used to pre-train robotic RL agents.

OpenAI Gym~\cite{brockman2016openai} has provided an elegant ecosystem and an abstraction layer between the learning algorithms and the environments. Currently OpenAI gym supports classical control tasks and such environments as Atari, MuJoCo, Box2D and OpenAI robotics environments based on MuJoCo that support simulated creatures, Fetch research platform and Shadow Dexterous Hand\textsuperscript{\textsc{tm}}. OpenAI Gym was created to provide a benchmarking platform for RL research by introducing strict naming and versioning conventions, making it possible to compare the results achieved by different algorithms and track the progress in the field.

Zamora et al. \cite{zamora2016extending} introduced an interface to integrate the Gazebo robotics simulator with the OpenAI Gym ecosystem, allowing to extend the set of possible RL environments to any that can be simulated in Gazebo. In their recent work, James et al.~\cite{james2019pyvrep} introduced a toolkit for robot learning research based on V-REP simulator. Another step in this direction is the PyRobot project~\cite{murali2019pyrobot} that provides a high-level interface for control of different robots via the Robot Operating System (ROS).

Although these tools provide an easy access to a variety of environments with the focus on specific tasks, all of these publicly accessible environments are still limited to simulation, only tangentially addressing the challenge of creating intelligent agents in the real physical world. The very few projects that have provided physical systems for community-driven robotics research are the LAGR~\cite{jackel2006darpa} project from DARPA, Georgia Tech's Robotarium~\cite{pickem2017robotarium} and TeleWorkBench~\cite{tanoto2006teleworkbench} from Bielefeld University. While being the closest to the concept of \iftoggle{final}{OffWorld}{(Anonymized)} Gym, the LAGR program has concluded and is not active anymore. TeleWorkBench and Robotarium did not postulate a specific task and thus do not serve as a benchmark challenge. Robotarium's maximum script execution time of 600 seconds makes it unsuitable for RL research. Moreover, none of the previous systems provided close integration into modern RL research ecosystem, proposed specific and version-controlled challenges nor had the same level of public accessibility as \iftoggle{final}{OffWorld}{(Anonymized)} Gym.

%
%  OffWorld Gym
%
\section{\iftoggle{final}{OFFWORLD}{(ANONYMIZED)} GYM}
\label{sec:gym}
\iftoggle{final}{OffWorld}{(Anonymized)} Gym is a framework with the goal of enabling the machine learning community to advance reinforcement learning for real-world robotics by validating and comparing different learning methods on a collection of real-world tasks. The framework consists of real-world environments in physical enclosures and their simulated replicas along with the necessary hardware and software infrastructure to access and run the experiments. There are four environments currently implemented in \iftoggle{final}{OffWorld}{(Anonymized)} Gym collection and presented in this work.

The first pair of environments feature a navigation task in a walled enclosure in which a wheeled robot has to traverse an uneven Moon-like terrain to reach an alluring visual beacon introduced by Kubrick et al.~\cite{kubrick19682001}. The robot receives $320 \times 240$ visual input from an RGBD camera and nothing else. The \texttt{MonolithDiscreteReal} environment features a discrete action space with four actions: left, right, forward, backward, each applying a velocity to the robot with a 2-second step duration. The continuous action space variant, \texttt{MonolithContinuousReal,} alternatively provides smooth controls to the linear and angular velocities of the robot. A sparse reward of $+1.0$ is assigned when the robot (Husarion Rosbot~\cite{rosbot2019manual}, dimensions $20.0 \times 23.5 \times 22.0$ cm) approaches the monolith within a radius of $40.0$ cm. The environment is reset upon successful completion of the task, reaching the limit of $100$ steps or approaching the boundary of the environment. After each reset, the robot is moved to a random position and into a random orientation. Figure \ref{fig:sim-real-env} (left) shows the environment and the input stream that the robot receives.

The second pair of environments inherits all of the characteristics of the first one, but is made more challenging by a different enclosure, which features obstacles that the robot has to avoid (see Figure \ref{fig:real-obstacles}). Both discrete and continuous action space environments are defined in a similar to manner to the first pair: \texttt{MonolithObstaclesDiscreteReal} and \texttt{MonolithObstaclesContinuousReal}. Developing a robust solution for this task would demonstrate the applicability of reinforcement learning approach to the problem of visual obstacle avoidance in absence of a map and location information. 

%\begin{figure}[h]
%    \caption{The real environment with obstacles and a sparse reward for reaching the monolith. An agent has to solve the problem of visual obstacle avoidance to complete the task successfully.}
%    \label{fig:real-obstacles}
%\end{figure}

As we further expand the \iftoggle{final}{OffWorld}{(Anonymized)} Gym framework's collection by building additional enclosures with various robotic tasks, we will cover a wide range of challenges for robotic systems, provide stable benchmarks, and make a step toward applicability of developed solutions to real world and industrial applications.

\subsection{Physical characteristics of an environment}
\label{sec:real-gym}
\begin{figure}[h]
    \centering
    \begin{floatrow}
        
        \ffigbox[6.12cm]{\caption{A real environment with obstacles and a sparse reward for reaching the monolith. An agent has to solve the problem of visual obstacle avoidance to complete the task successfully.}\label{fig:real-obstacles}}{
        \includegraphics[width=1.0\linewidth]{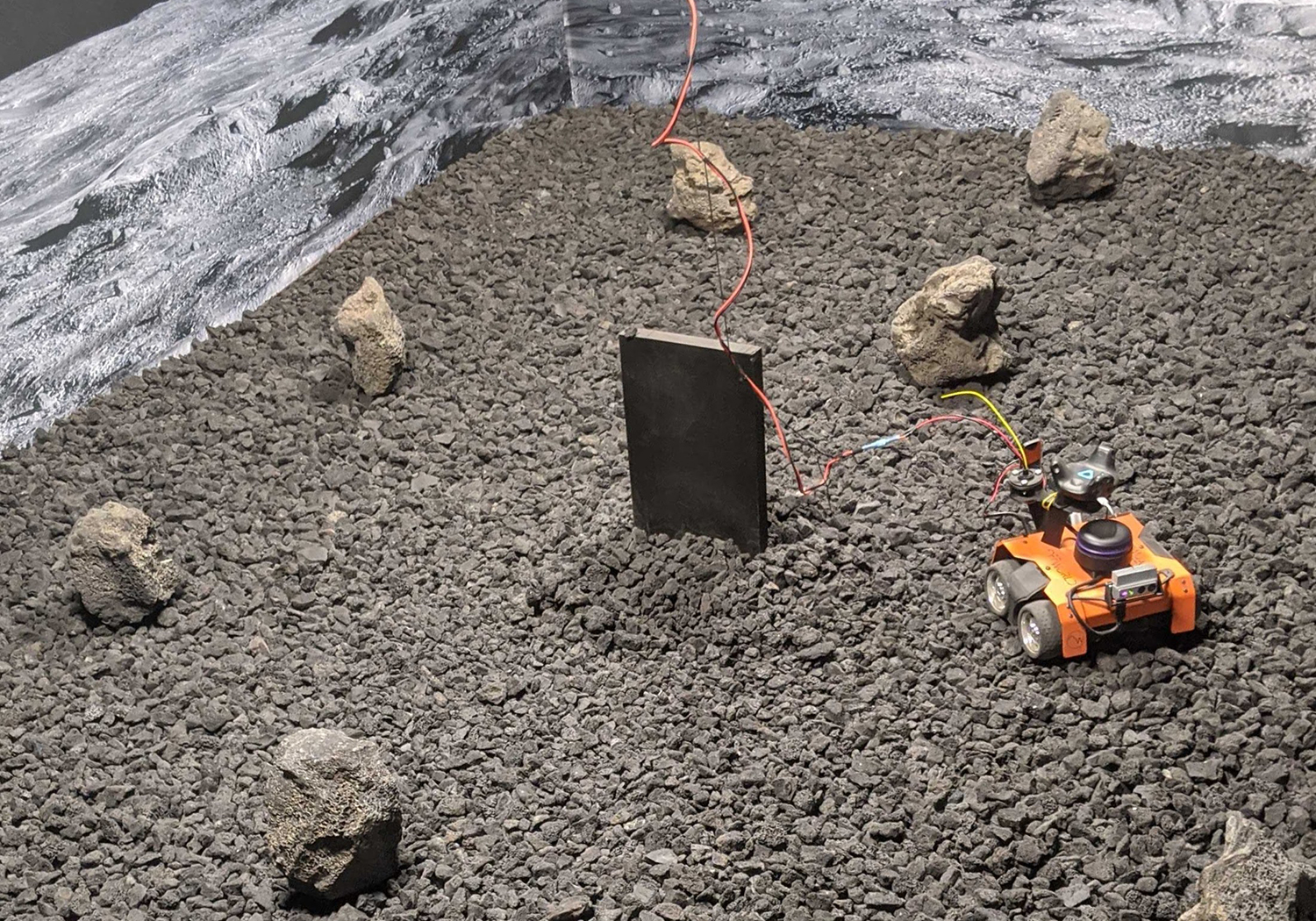}}
    
        \ffigbox[7.48cm]{\caption{Internal representation of the real environment by the environment control system. Two \texttt{lighthouse} components are tracking the position of the \texttt{tracker} that is attached to the \texttt{base\_link} of the robot. The monolith is installed in the middle of the world coordinate frame. The yellow line shows a global plan created by the \texttt{move\_base} for resetting an episode, and the red arrow indicates the desired final orientation of the robot.}\label{fig:rviz-setup}}{
        \includegraphics[width=1.0\linewidth]{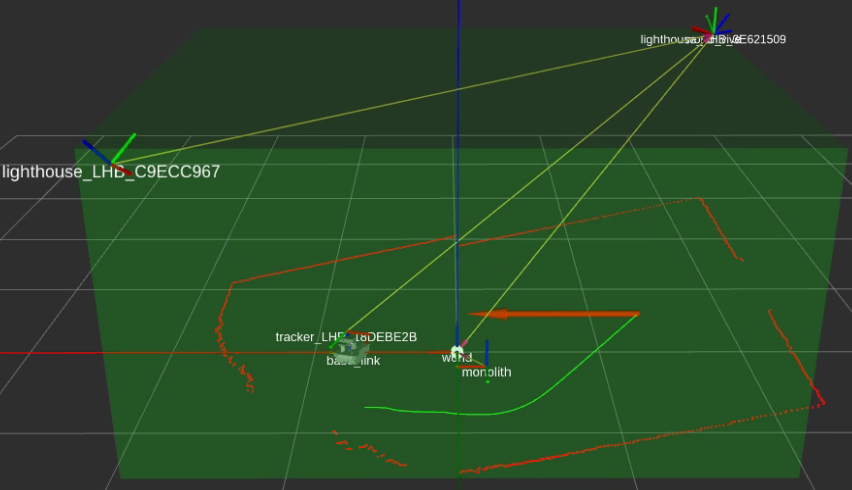}}
    
    \end{floatrow}
\end{figure}

A real instance of an environment is an enclosure of size $3 \times 4 \times 2$ meters designed to visually emulate the lunar surface. The ground layer is covered with small lava rocks that create an uneven terrain that is challenging for the robot to traverse and prevents the robot from having stable visual observations. The enclosure provides power to the robot, network connection to the server that is running the environment, and two overhead cameras that allow the user to monitor the environment remotely. An HTC Vive\textsuperscript{\textsc{tm}} tracker and two base stations are used to localize the robot within the environment. Localization information is not available to the learning agent but is used internally by the environment control script to calculate rewards, reset the environment and achieve new initial locations at the start of each episode. Figure \ref{fig:rviz-setup} shows the internal representation of the real environment used by the \iftoggle{final}{OffWorld}{(Anonymized)} Gym server to control and monitor the environment. 

%\begin{figure}[h]
%    \includegraphics[width=1.0\linewidth]{images/rviz-setup.png}
%    \caption{Internal representation of the real environment by the environment control system. Two \texttt{lighthouse} components are tracking the position of the \texttt{tracker} that is attached to the \texttt{base\_link} of the robot. The monolith is installed in the middle of the world coordinate frame. The yellow line shows a global plan created by the \texttt{move\_base} for resetting an episode, and the red arrow indicates the desired final orientation of the robot.}
%    \label{fig:rviz-setup}
%\end{figure}

%We define a 3D transformation matrix to allow transformation from the tracker's coordinate frame to the world coordinate frame (defined as the center of geometry of the enclosure), and another 3D transformation matrix for transformation from the tracker's coordinate frame to the robot's coordinate frame. These transformation matrices help determine the robot's location with respect to the world's coordinate frame at any time during the experiment. Upon an environment reset the robot is moved to a randomly chosen spawn location using localization information from the HTC Vive setup and motion control using ROS's \texttt{move\_base} navigation and path planning package. 

\subsection{Simulated analog environments}
\label{sec:sim-gym}
The alternate simulated variants of each of the \iftoggle{final}{OffWorld}{(Anonymized)} Gym environments are created using Gazebo simulation software and provide a close replica of the physical environments. In addition to the default applications of simulated environments, such as algorithm development and preliminary testing of the agent, the close match between the \iftoggle{final}{OffWorld}{(Anonymized)} Gym simulated and real instances provides a platform for researching feasibility and methodology of transferring agents that are trained in simulation to the real world. Simulated environments imitate the dimensions, physical parameters of the real system such as mass and friction of the robot, reward and reset criteria, and the visual appearance as close to the real environment as possible. To make the simulation simpler and more resource-efficient, the ground is modeled as a single uneven mesh and not a collection of distinct rocks. For users wanting a simple simulation environment installation, versions of the simulation Gym environments are available wrapped inside of Docker containers and interfaceable from the user's host machine.

\subsection{The architecture of the system}
\label{sec:gym-implementation}

\begin{figure}[h]
    \includegraphics[width=1.0\linewidth]{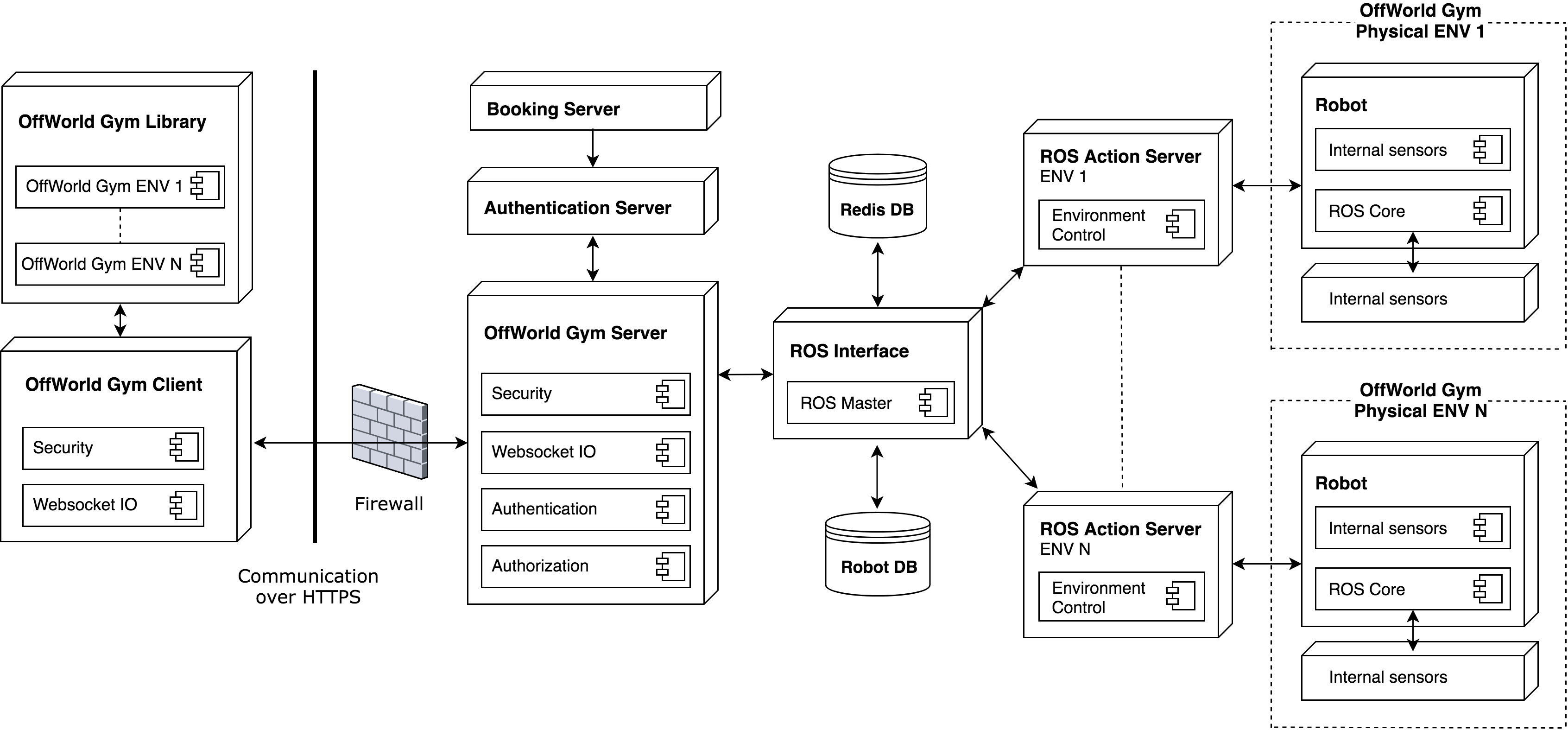}
    \caption{System architecture of \iftoggle{final}{OffWorld}{(Anonymized)} Gym}
    \label{fig:architecture}
\end{figure}

\iftoggle{final}{OffWorld}{(Anonymized)} Gym consists of three major parts: (a) a Python library that is running on the client machine, (b) the server that handles communication, resource management and control of the environment (reward, episode reset, etc.), (c) the physical enclosure that provides power and network infrastructure, and (d) the robot itself. Figure \ref{fig:architecture} gives an overview of the architecture, its components and interactions. 

The \iftoggle{final}{OffWorld}{(Anonymized)} Gym library provides the API to access the environments. The client side of the library accepts actions and commands from the user's RL code and forwards them to the gym server. The gym server controls the resource management and, if the client has access, transforms the request into a sequence of ROS requests, which are then forwarded to the ROS action server controlling the physical environment. The ROS action server validates each command and forwards it to the robot. Physical execution of an action by the robot can take up to 4 seconds: 2 seconds are taken by the fixed-length step duration and the rest of the time varies depending on the load of the server, network latency and the time of the code execution. The robot completes the requested action (movement, position reset, etc) and sends the final telemetry readings back to the action server. The server pre-processes the telemetry and creates the state variable that is sent back to the client as an observation for the agent. The user does not have a direct access to the robot and can only communicate via the established set of telemetry messages and control commands. The control logic and the learning process are executed on user's workstation and the user is thus free to explore any algorithmic solutions and make use of any amount of computational resources available at their disposal.

We have closely followed the ecosystem established by OpenAI Gym so that the deployment of an agent in our environment requires minimal change when switching from any other \texttt{gym} environment. Listing 1 illustrates the conceptual blocks of the program that uses our environment to train a reinforcement learning agent.

\begin{python}[From the user perspective switching to \iftoggle{final}{OffWorld}{(Anonymized)} Gym is done by simply changing the name of the environment.]
import gym
import offworld_gym
from offworld_gym.envs.common.channels import Channels
from rl.agents.dqn import DQNAgent
...
# connect to real environment
env = gym.make('OffWorldMonolithDiscreteReal-v0',
               experiment_name='My new experiment',
               resume_experiment=False,
               channel_type=Channels.DEPTH_ONLY)

# or connect to the simualted environment
env = gym.make('OffWorldMonolithDiscreteSim-v0',
               channel_type=Channels.DEPTHONLY)
...
model = create_network(...)
dqn = DQNAgent(model=model, ...)
dqn.compile(...)
dqn.fit(env, ...)
\end{python}

To deploy an agent in an \iftoggle{final}{OffWorld}{(Anonymized)} Gym environment a user has to install \iftoggle{final}{\texttt{offworld\_gym}}{\texttt{anonymized\_gym}} Python library and register with the resource management system\footnote{See \iftoggle{final}{\texttt{\url{https://gym.offworld.ai}}}{\texttt{\url{https://anonymized}} \emph{(please see screenshots of the critical components of the online system attached to this submission)}} for details.}. 

\subsection{Hardware specification}
\label{sec:gym-hardware}

The Husarion Rosbot is equipped with an ASUS Up Board (Quad Core Intel CPU, Ubuntu 16.04) on-board computer, Orbbec Astra RGBD camera and a CORE2-ROS robot controller. The robot controller runs the firmware layer and the on-board computer runs the sensor drivers, ROS sensor packages and robot motion controller ROS package. Since all of the learning happens on the client workstation, the on-board capabilities of the robot can be kept minimal. An Intel NUC (Core i7, 32 GB RAM, Ubuntu 16.04) computer runs the \iftoggle{final}{OffWorld}{(Anonymized)} Gym Server, the robot mission management software and the ROS packages that control the environment. An IBM workstation (Intel Xeon, 32 GB RAM, Nvidia Quadro, Ubuntu 16.04) interfaces with the HTC Vive lighthouse setup. It runs the HTC Vive driver and a ROS package which publishes the robot's localization data.

%
%  Experiments
%
\section{EXPERIMENTS}
\label{sec:result}
The purpose of our experimental work is threefold: to demonstrate the soundness of the system and feasibility of learning, provide the first set of benchmark results for modern RL algorithms, and to empirically estimate the sample complexity of learning a visual navigation task end-to-end on a real robot from camera inputs directly to actions.

\begin{figure}[h]
    \includegraphics[width=\linewidth]{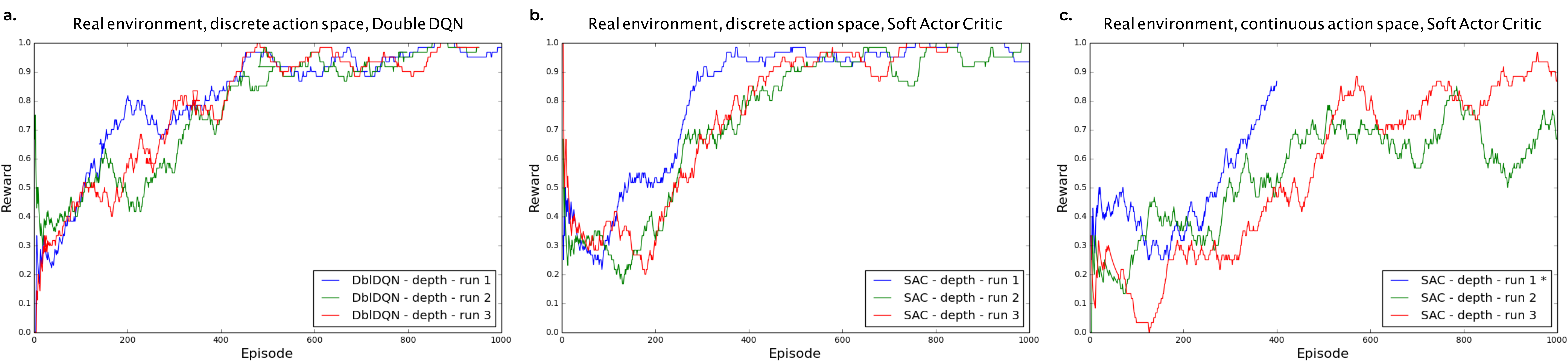}
    \caption{Learning curves in discrete and continuous variants of the environment. \textbf{a.} Double DQN trained end-to-end with discrete actions space: \iftoggle{final}{\url{https://www.youtube.com/watch?v=lgwaZHxtJc0}}{\texttt{\url{https://youtu.be/FkUJ_Gl1hJU}}} \textbf{b.} Soft Actor-Critic solves the discrete actions space variant of the environment. \textbf{c.} SAC achieves intelligent behavior in the continuous variant of the environment. %: \iftoggle{final}{INSERT PUBLIC VID}{\texttt{\url{https://youtu.be/VsHNX6WTi8Q}}}.
    }
    \label{fig:learning-curves}
\end{figure}

We trained Double DQN \cite{van2016deep} and Soft Actor-Critic (SAC)~\cite{haarnoja2018soft} agents in the discrete action space variant of \texttt{MonolithReal} environment and a SAC agent in the continuous variant of the same environment. Figure \ref{fig:learning-curves} shows the learning curves for all three experiments. We have also trained the same agents in simulated versions of the environments, we are not presenting these results here as we want to focus on learning in real. When deployed in simulated environments the same architectures achieved similar results with similar sample complexity.

The Double DQN agent's neural network architecture consisted of a $320 \times 240$ visual (depth channel only) input, followed by three convolutional layers each with four $5 \times 5$ stride-$2$ filters and max pooling, followed by two fully connected layers of size $16$. Leaky ReLU activations were used. In total the network had 3381 trainable parameters. The Adam optimizer was used with a learning rate of $0.001$, a batch size of $32$, and a target network Polyak update Tau of $0.01$. The circular replay buffer was of size $25,000$, and experience was gathered in an epsilon-greedy fashion, where epsilon was linearly annealed from $0.9$ to $0.1$ over the first $40,000$ steps. The discount factor was $0.95$.

The SAC agent's neural network architecture consisted of an $84 \times 84$ visual (depth channel only) input, followed by three convolutional layers with $16$, $32$, and $64$ filters of sizes $8 \times 8$, $4 \times 4$, and $1 \times 1$, and strides of $4$, $2$, and $1$, respectively. This was followed by two fully connected layers of size $64$. ReLU activations were used. In total the network had 757,236 trainable parameters. The Adam optimizer was used with a learning rate of $0.0003$, a batch size of $1024$, and a target network Polyak update Tau of $0.005$. Updates were performed after every experience step. The circular replay buffer was of size $500,000$. $\alpha$ was learned to match an entropy target of $0.2 * -\log(1/|A|)$ for discrete spaces and $0.2*-\dim(A)$ for continuous spaces. The discount factor was $0.99$.

The results confirm the overall soundness of the proposed system and demonstrate feasibility of learning. We count on community involvement to evaluate the other existing algorithms, explore different architectures and methods in order to identify the state of the art algorithms for the tasks presented in \iftoggle{final}{OffWorld}{(Anonymized)} Gym. To this end we provide open remote access to the environments and encourage sharing of the results achieved by different methods via the Leaderboard\footnote{\iftoggle{final}{\texttt{\url{https://gym.offworld.ai/leaderboard}}}{\texttt{\url{https://anonymized/leaderboard}}}}, a component of our system that allows to log and compare the performance of different approaches.

%
%  Conclusion
%
\section{CONCLUSION}
\label{sec:conclusion}
In this work we present a collection of real-world environments for reinforcement research in robotics. We aim to build up a standard real-world benchmark for RL research, that allows to test learning algorithms not only in simulated and game environments, but also on real robots and real-world tasks.

Working with real physical environments pose significant challenges for the speed of progress in RL research. Inability to run the experiments faster than real time, mechanical difficulties with the robots and supporting mechanisms, unpredictable behavior of the real physical medium, the cost of the system, and the additional time for resetting the environment between episodes are major technical challenges that have slowed down the advancement of RL in real robotics. Furthermore, in a simulated environment we can engineer any reward schema required by the experimental setup, whereas in the real world reward specification is limited by the sensors a robot has and their robustness. Despite all these challenges, the alternative -- robotic simulation -- can only partially address all the relevant aspects of real robotic behavior. For the real deployment of RL systems the community will have to face the above-mentioned challenges. We hope that the interaction with \iftoggle{final}{OffWorld}{(Anonymized)} Gym will provide valuable insights into these challenges and facilitate the search for solutions to them.

The \iftoggle{final}{OffWorld}{(Anonymized)} corporation is committed to providing long-term support of \iftoggle{final}{OffWorld}{(Anonymized)} Gym environments to ensure that they can serve as a benchmark for RL research. By taking care of the maintenance of both the hardware and software components of the system, as well as construction of additional environments, \iftoggle{final}{OffWorld}{(Anonymized)} ensures that RL community can focus on the algorithmic side of the challenge and not spend time on the challenges posed by the hardware.

The \iftoggle{final}{OffWorld}{(Anonymized)} Gym architecture has been designed to abstract away the complexities and particularities of handling a physical robot system from the user. Close integration into existing ecosystem of OpenAI Gym allows to use the environment without any prior experience in robotics, abstracting it under a familiar API and taking off the burden of hardware cost and maintenance. The scalability of the system is addressed by monitoring user activity via the time booking system and building additional physical environments to meet the demand.

We also provide simulated environments that are close replicas of the real ones as part of the same framework. This allows to setup and validate experiments in simulation ahead of real deployment, experiment with learning techniques that rely on pre-training in simulation, domain adaptation to close the reality gap, domain randomization and other techniques that help reduce sample complexity of RL in the real world.

The experiments in training Double DQN and SAC agents in the proposed environment confirm the soundness of the system and show feasibility of learning. They also provide initial benchmark results that we hope will soon be surpassed by the novel algorithms and approached proposed by the reinforcement learning and robotics community.

Our future work includes building and releasing more enclosures with various tasks. We aim to maintain a focus on industrial robotic challenges in unstructured environments, striving towards general applicability of the methodologies that will be discovered in these environments to real-world applications. Future work also includes benchmarking other existing RL algorithms, imitation learning methods, and transfer of the agents trained in simulation to real environments. This research will show which methods are the most efficient in terms of sample complexity, optimality and robustness of achieved behaviours and their resilience to the different kinds of stochasticity (environment, sensory, reward, action) a real environment can present.

%
%  Acknowledgements
%
\acknowledgments{Special thanks to Eric Tola, Matt Tomlinson, Matthew Schwab and Piyush Patil for help with the mechanical and electrical design and implementation.}

%
%  References
%
\setlength{\bibsep}{1.5pt}
{\footnotesize
\bibliography{main}}

\end{document}